\documentclass[conference]{IEEEtran}
\IEEEoverridecommandlockouts
\usepackage{cite}
\usepackage{amsmath,amssymb,amsfonts}
\usepackage{algorithmic}
\usepackage{graphicx}
\usepackage{pdflscape}
\usepackage{textcomp}
\usepackage{xcolor}
\usepackage{rotating}
\def\BibTeX{{\rm B\kern-.05em{\sc i\kern-.025em b}\kern-.08em
    T\kern-.1667em\lower.7ex\hbox{E}\kern-.125emX}}
\begin{document}

\title{An Underexplored Frontier: Large Language Models for Rare Disease Patient Education and Communication -- A scoping review\\
\thanks{%
\textsuperscript{$^\dagger$} Equally contributions.

\textsuperscript{*} Corresponding authors.

This work was supported by the U.S. National Science Foundation and the French National Research Agency joint program, project CIL-FLARE (NSF award number 2530226, ANR grant number ANR-25-CE45-4963). }
}

\author{

\IEEEauthorblockN{
    Zaifu Zhan\textsuperscript{1,2,$^\dagger$},
    Yu Hou\textsuperscript{1},
    Kai Yu\textsuperscript{1},
    Min Zeng\textsuperscript{1},
    Anita Burgun\textsuperscript{3,4},
    Xiaoyi Chen\textsuperscript{1,3,$^\dagger$,*},
    Rui Zhang\textsuperscript{1,*}
}

\IEEEauthorblockA{\textsuperscript{1} \textit{Division of Computational Health Sciences, Department of Surgery, University of Minnesota}, Minneapolis, USA}

\IEEEauthorblockA{\textsuperscript{2} \textit{Department of Electrical and Computer Engineering, University of Minnesota}, Minneapolis, USA}

\IEEEauthorblockA{\textsuperscript{3} \textit{Clinical Bioinformatics Lab, Université Paris Cité, Institut Imagine, INSERM UMR 1163}, Paris, France}

\IEEEauthorblockA{\textsuperscript{4} \textit{Medical Informatics Department, Hôpital Necker-Enfants Malades, AP-HP}, Paris, France}


\IEEEauthorblockA{\{zhan8023, hou00127, yu001014, minzeng, chen9435, ruizhang\}@umn.edu, anita.burgun@aphp.fr}
}

\maketitle

\begin{abstract}
Rare diseases affect over 300 million people worldwide and are characterized by complex care pathways, limited clinical expertise, and substantial unmet communication needs throughout the long patient journey. Recent advances in large language models (LLMs) offer new opportunities to support patient education and communication, yet their application in rare diseases remains unclear.

We conducted a scoping review of studies published between January 2022 and March 2026 across major databases, identifying 12 studies on LLM-based rare disease patient education and communication. Data were extracted on study characteristics, application scenarios, model usage, and evaluation methods, and synthesized using descriptive and qualitative analyses.

The literature is highly recent and dominated by general-purpose models, particularly ChatGPT. Most studies focus on patient question answering using curated question sets, with limited use of real-world data or longitudinal communication scenarios. Evaluations are primarily centered on accuracy, with limited attention to patient-centered dimensions such as readability, empathy, and communication quality. Multilingual communication is rarely addressed.

Overall, the field remains at an early stage. Future research should prioritize patient-centered design, domain-adapted methods, and real-world deployment to support safe, adaptive, and effective communication in rare diseases.
\end{abstract}

\section{Introduction}
Rare diseases, defined in Europe as conditions affecting fewer than 1 in 2,000 individuals and in the United States as those affecting fewer than 200,000 people, collectively represent a major global health challenge. Despite their individual rarity, more than 7,000 rare diseases collectively affect an estimated 300 million people worldwide\cite{lancet2024rare}. About 80\% of rare diseases have a genetic origin, and more than 90\% lack approved treatments\cite{1002021100}.

Rare disease patient communication is uniquely complex and differs fundamentally from that in more common conditions. Patients and families often move through a long diagnostic odyssey\cite{bauskis2022diagnostic}, interact with many clinicians who have limited condition-specific expertise\cite{gunes2025challenges}, and must communicate across uncertainty\cite{devisetti2024embracing}, fragmented care\cite{balfour2026patient,groza2026reimagining}, and information scarcity\cite{mascalzoni2014rare}. Communication is particularly critical at the diagnostic stage, which remains one of the most challenging phases in rare diseases research and care. Delivering a rare disease diagnosis requires not only explaining what is known and unknown about etiology, prognosis, and treatment, but also addressing prior harms from delayed or missed diagnoses and rebuilding trust\cite{merker2021effective}. Beyond diagnosis, communication needs extend across the patient journey. In the absence of effective treatments for most rare diseases, daily self-management becomes central to maintaining quality of life, creating ongoing needs for communication around evolving symptoms, functional limitations, fatigue, nutrition, work, and family planning, often under significant cognitive and emotional burden. However, these long-term communication needs remain relatively  underexplored\cite{cribbs2025capturing}. Across these stages, the multidimensional complexity of rare diseases further amplifies health literacy challenges, as patients must interpret complex, uncertain, evolving, and sometime inconsistent information, spanning diagnosis, prognosis, treatment and daily life. Together, these factors highlight the need for rare disease communication to be not only accurate, but also adaptive, longitudinal, and responsive to patient-specific contexts. 

Existing research highlights several persistent gaps in rare disease patient communication. Studies consistently report difficulties in accessing understandable and trustworthy information and limited support for shared decision-making, and insufficient attention to family- and caregiver-centered communication\cite{stenberg2024scoping,baynam2024global}. Despite the recognized complexity of communication across the rare disease journey, the current evidence base remains largely descriptive, with a predominance of qualitative studies documenting patient experiences. There is a lack of communication approaches capable of addressing the diversity, heterogeneity, and evolving needs of rare disease populations, particularly in real-world settings, highlighting the need for scalable, adaptive, and patient-centered solutions.

\begin{figure}[t]
    \centering
    \includegraphics[width=0.99\linewidth]{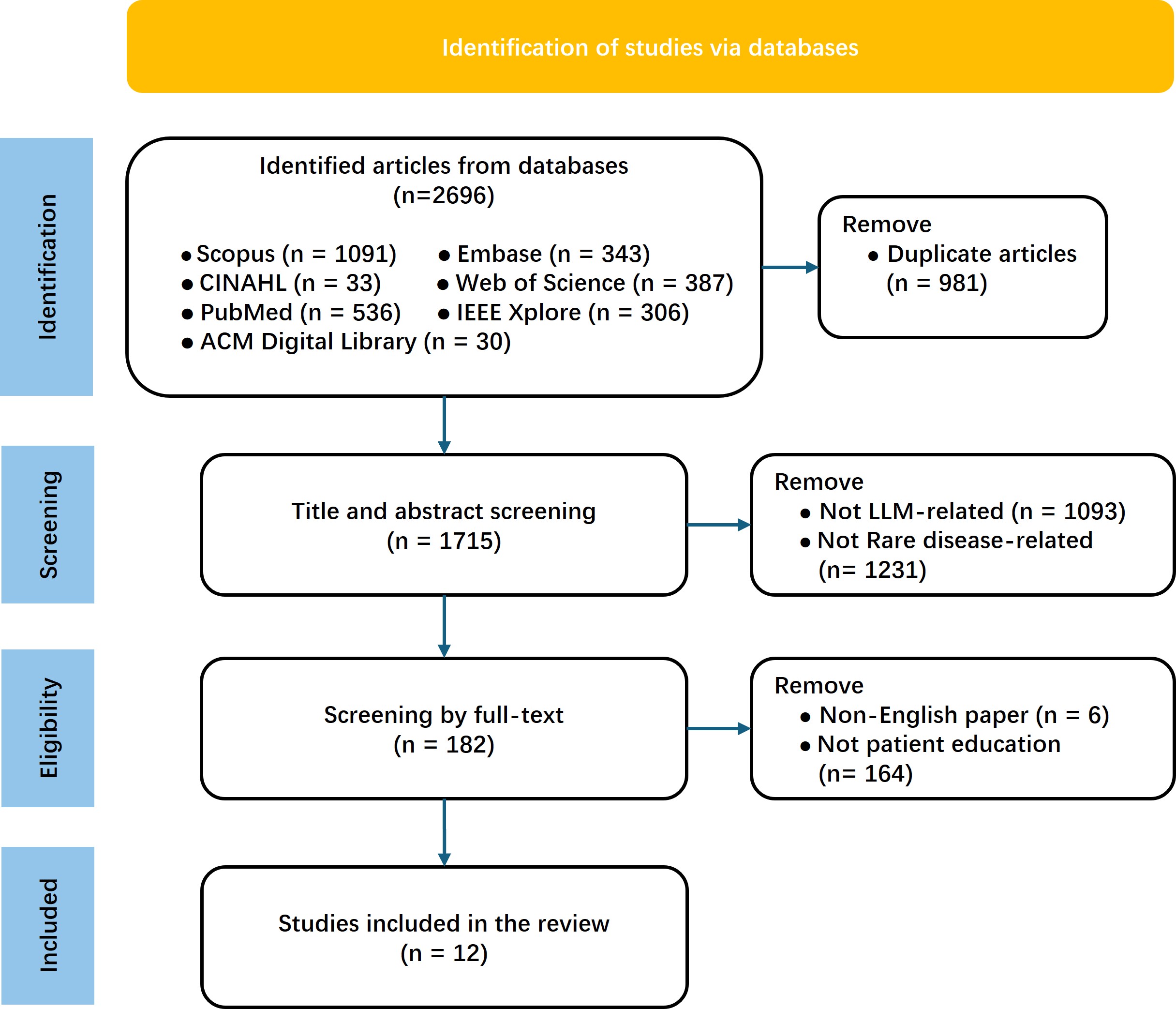}
    \caption{PRISMA flowchart of study records.}
    \label{fig:method}
\end{figure}
In this context, recent advances in large language models (LLMs)\cite{zhou2025large} offer new opportunities to rethink patient communication in rare diseases. LLMs have demonstrated strong capabilities in natural language understanding and generation, enabling them to process unstructured clinical notes\cite{zhan2025ramie}, patient narratives\cite{ghenimi2025paradox}, and biomedical literature\cite{hou2025benchmarking}, and to generate context-aware, conversational responses. Recent models such as GPT-5 demonstrate remarkable capacity for answering complex questions, synthesizing information from heterogeneous sources\cite{hou2025benchmarking}. 
Beyond these general capabilities, LLMs are particularly well-suited to address the specific challenges of rare disease communication. Their ability to integrate fragmented and sparse knowledge may help bridge gaps in clinician expertise and information availability\cite{zeng2026medclbench}. Interactive dialogue capabilities enable iterative clarification, uncertainty communication, and adaptation to patient knowledge levels, which are critical in contexts where patients often act as “expert patients.” In addition, LLMs offer the potential to incorporate patient-generated data, such as symptom descriptions and lived experiences, to support more personalized and context-aware communication across the patient journey. Compared to traditional static educational materials or rule-based systems, LLMs can support dynamic, longitudinal, and patient-centered interactions.
Unlike traditional static educational materials, LLM-based systems can support interactive, adaptive, and personalized communication, potentially bridging gaps between clinical knowledge and patient-specific concerns. In rare diseases specifically, where specialist knowledge is scarce and geographically concentrated, such tools provide particular potential to reduce information inequity and support patient empowerment. 
At the same time, their application in rare diseases raises important challenges. Ensuring factual accuracy is particularly critical in a domain characterized by limited evidence and high uncertainty. Risks related to hallucination, outdated or incomplete knowledge, and lack of transparency may have significant consequences for patients. Furthermore, designing systems that appropriately balance information delivery with empathy, respect patient expertise, and support shared decision-making remains an open research question. These considerations highlight both the promise and the need for careful evaluation of LLM-based approaches in rare disease patient communication.
Recently, a growing body of research has begun to explore the application of LLMs to rare disease patient communication and education. This review aims to provide an overview of current applications, their approaches, evaluation strategies and limitations, and identify future directions for the development of safe, effective, and patient-centered LLM-based communication systems in rare diseases.

\section{Methods}\label{sec:method}

\subsection{Literature search strategy}

We conducted a systematic literature search to identify studies on the application of large language models (LLMs) in rare disease patient communication and education. The search was performed across multiple electronic databases, including PubMed, Web of Science, IEEE Xplore, Scopus, Embase, CINAHL, and ACM Digital Library, covering publications from January 2022 to early 2026.

The search strategy was constructed using Boolean logic to combine three key concept groups: (1) LLM-related terms, (2) rare disease-related terms, and (3) disease-related descriptors. Specifically, the query was defined as:

\begin{verbatim}
("llm*" OR "language model*" OR 
"foundation model*" OR "GPT*" OR
"generative")
AND
("rare*" OR "orphan*" OR "mendelian*" OR 
"monogenic*" OR "inherited*" OR "genetic*")
AND
("disease*" OR "disorder*")
\end{verbatim}

Wildcard operators (*) were used to capture variations of key terms. The search syntax was adapted as needed for different databases. To ensure comprehensive coverage, we additionally screened the reference lists of relevant articles and recent review papers. The final search was conducted in early 2026.

\subsection{Eligibility criteria}

Studies were included if they met the following criteria:
\begin{enumerate}
    \item Focused on the use of LLMs or LLM-based systems;
    \item Addressed rare diseases or included rare disease scenarios;
    \item Involved patient-facing applications, such as question answering, education, counseling, or communication support;
    \item Reported empirical results, evaluation metrics, or case analyses.
\end{enumerate}

\subsection{Study selection process}

As shown in Fig. \ref{fig:method}, all retrieved records were first screened based on titles and abstracts, with the help of GPT-4o\cite{achiam2023gpt}. Full-text reviews were conducted for potentially relevant studies. The selection process followed a structured workflow to ensure consistency and reproducibility.

Disagreements regarding study inclusion were resolved through discussion among the authors until consensus was reached.

\subsection{Data extraction}

For each included study, we extracted key information across multiple dimensions, including:
\begin{itemize}
    \item Publication information (year, region);
    \item Rare Disease (and category);
    \item Application scenario (e.g., question answering, education generation, counseling);
    \item Data sources (e.g., patient questions, clinical text, social media);
    \item LLM models used;
    \item Evaluation methods;
    \item Key contributions, limitations, and future directions.
\end{itemize}

The extracted data were organized into a structured dataset to enable cross-study comparison and synthesis.

\subsection{Data analysis}
Descriptive statistics were used to summarize study characteristics across the defined dimensions. Frequencies and distributions were calculated for categorical variables such as disease types, model usage, and evaluation methods.

In addition, qualitative synthesis was conducted to identify common patterns in study contributions, limitations, and emerging research directions.


\begin{sidewaystable*}
\caption{Large Language Models in Rare Disease Education - Comprehensive Dataset (n=12). Yr=Year, Q=Question(s), Cat.=Category, Eval=Evaluation Type, CV=Cardiovascular, Autoinflam=Autoinflammatory, NDD=Neurodevelopmental Disorder, Epi=Epilepsy}
\label{tab: papers}
\centering
\small
\begin{tabular}{cp{2.2cm}cp{0.8cm}p{1.2cm}p{1.8cm}p{0.95cm}p{1.5cm}p{0.95cm}p{2.1cm}p{1.8cm}p{1.8cm}p{1.8cm}}
 
\hline
\textbf{No.} & \textbf{Short Title} & \textbf{Task} & \textbf{Data} & \textbf{Region} & \textbf{Disease(s)} & \textbf{Cat.} & \textbf{LLM(s)} & \textbf{Eval} & \textbf{Metric} & \textbf{Contribution} & \textbf{Limitation} & \textbf{Future Work} \\
 
\hline
 
1 & Can ChatGPT Answer Patient's Questions?\cite{tao2025can} & QA & 1,467 Q\&A & USA & Rare genetic & Genetic & GPT-4.0 & Expert & Accuracy, comprehensiveness & High accuracy on genetic Qs & Inaccuracies require verification & Add clarity/relevance criteria \\
 
 
 
\hline
 
2 & ChatGPT's Cardiovascular Education\cite{sevincc2025academic} & QA & 40 Q & Turkey, Germany, Australia & 7 rare CVDs & CV & ChatGPT & Expert & Accuracy & Reasonable CV education quality & Less reliable for rare vs. common & Improve patient-phrased responses \\
 
\hline
 
3 & AI Chatbots and Narcolepsy\cite{henriques2025artificial} & QA & 28 Q & Portugal & Narcolepsy & Neuro & ChatGPT, Gemini, Perplexity & Expert & Accuracy, Completeness, Readability & Accurate, complete narcolepsy info & High reading level; accessibility issues & Improve readability \& professional integration \\
 
\hline
 
4 & ChatGPT Trustworthiness for Sarcoma\cite{valentini2024artificial} & QA & 25 Q & Austria & Sarcoma & Cancer & ChatGPT & Expert & Accuracy & Variable quality with 3 experts & GPT-3.5 only; small sample & Supervised AI use with oncologists \\
 
\hline
 
5 & LLMs in Dermatologic Education\cite{lambert2024assessing} & Generation & 960 materials & USA & 8 derm conditions (4 common, 4 rare) & Derma & ChatGPT-3.5, GPT-4, DocsGPT, DermGPT & Expert & Accuracy, Reading level & Appropriate reading level generation & Small sample; potential bias & Multilingual exploration; assess trust \\
 
\hline
 
6 & Automating LLM Response Evaluation\cite{zhao2025automating} & QA & 175 responses to 25 Q & USA & Complex Lymphatic Anomalies & Vascular & GPT-4, GPT-4o, Qwen, DeepSeek, Gemma, LLaMA & Expert + NLP & Accuracy & Compare NLP vs. LLM metrics & CLA-specific; limited generalizability & Test across diverse diseases \\
 
\hline
 
7 & ChatGPT, Gemini, Grok on FMF\cite{cilli2026chatgpt} & QA & 49 Q & Turkey & Familial Mediterranean Fever & Autoinflam & ChatGPT, Gemini, Grok & Expert & Accuracy & Variable accuracy across models & English only; no multilingual test & Multilingual validation; assess comprehension \\
 
\hline
 
8 & AI-Driven Podcast Generation\cite{perez2025enhancing} & Generation & 8 podcasts & USA, Europe & 6 rare epilepsies/NDD & Neuro & NotebookLM & Expert + stakeholder & Educational value, Accuracy & Scalable multilingual tool (EN, ES, DE) & Needs quality control; oversimplification & Enhance accuracy \& cultural sensitivity \\
 
\hline
 
9 & Generative AI for Genetic Counseling\cite{jeon2025evaluating} & QA & 102 Q & South Korea & 4 rare genetic diseases & mostly neuro & ChatGPT, Gemini, Claude, Perplexity & Expert & Accuracy, Professionalism & ChatGPT best; counseling potential & Occasional inaccuracies; low sources & Expert verification protocols \\
 
\hline
 
10 & LLMs in Fibromuscular Dysplasia\cite{tefera2025large} & QA & 10 Q & USA & Fibromuscular Dysplasia & Vascular & ChatGPT & Expert & Accuracy & No significant difference vs. physicians & Training data dependent; subjective & Safety assessments; rare disease training \\
 
\hline
 
11 & MedBot vs RealDoc\cite{weber2025medbot} & QA & 103 Q\&A & Germany/ English & 6 rare respiratory diseases & Respiratory & GPT-4, BioMistral 7B & Expert & Correctness, Comprehensibility, Relevance, Empathy & GPT-4 outperforms physicians & Small evaluator sample; language gaps & Test across languages; live content tuning \\
 
\hline
 
12 & ChatGPT for Rare Kidney (ERKNet)\cite{van2025risks} & QA & 54 experts & 13 EU countries & 28 rare kidney diseases & Kidney & ChatGPT-3.5, GPT-4.0 & Expert & Correctness, Helpfulness & International consensus on utility & Safety concerns; hallucinations possible & Curated dataset fine-tuning; protocols \\
 
 
 
\hline
 
\end{tabular}

\end{sidewaystable*}


\section{Results}
Based on the search strategies and methods described in Section \ref{sec:method}, we ultimately identified 12 papers to include in our review (Table. \ref{tab: papers}). In this section, we will examine these 12 papers from different perspectives (Fig. \ref{fig:res}).

\paragraph{Temporal Trends.}
The distribution of included studies reveals a clear temporal concentration in recent years. As shown in Fig. \ref{fig:res}(a), among the 12 studies analyzed, the majority were published in 2025 (n = 10), compared to a smaller number in 2024 (n = 2). This trend reflects the rapidly growing interest in applying large language models (LLMs) to healthcare, and more specifically, to patient-facing applications. The sharp increase in publications within a short time frame highlights the emerging and still-evolving nature of this research area, particularly in the context of rare diseases and patient education.

\paragraph{Disease Coverage.}
The included studies span a diverse range of disease domains, As shown in Fig. \ref{fig:res}(d), with neurological conditions (n = 3) representing the most frequently studied category at the paper level, followed by cardiovascular and respiratory diseases (n = 2 each). Other domains, including renal, dermatological, oncological, and lymphatic diseases, are represented by a smaller number of studies. Additionally, two studies address rare diseases in a more general or unspecified manner.
\begin{figure*}[t]
    \centering
    \includegraphics[width=0.99\linewidth]{results.jpg}
    \caption{Metadata of information from LLM-based rare disease patient communication and education studies included in this review.
    (a) Distribution of publications by year, showing a strong temporal concentration in 2024 and 2025, reflecting the recent emergence of this research area. 
(b) Distribution of LLM usage, where GPT-based models dominate across studies. 
(c) Distribution of task categories, indicating that question answering (QA) is the most common application scenario. 
(d) Distribution of rare disease categories at the study level. 
(e) Frequency of disease mentions across all included studies, providing a finer-grained view of disease coverage. 
(f) Distribution of evaluation metrics, with accuracy and correctness being the most frequently used measures. 
(g) Distribution of data source types, highlighting the predominance of patient-oriented question datasets. 
(h) Distribution of application scenarios, illustrating the focus on patient-facing communication tasks.
Overall, these results reveal a rapidly evolving field characterized by heavy reliance on GPT-family models, a strong emphasis on question answering tasks, and a predominant focus on accuracy-based evaluation, while more diverse data sources and patient-centered evaluation dimensions remain underexplored.
}
    \label{fig:res}
\end{figure*}
When examining disease mentions at a finer granularity, a broader distribution emerges Fig. \ref{fig:res}(e). Renal (n = 11) and neurological (n = 10) conditions appear most frequently across all studies, followed by cardiovascular (n = 8) and respiratory diseases (n = 6). This discrepancy suggests that while individual studies may focus on specific domains, many incorporate multiple disease contexts, indicating an effort to evaluate LLMs across heterogeneous clinical scenarios. However, certain disease categories remain underrepresented, highlighting potential gaps in current research coverage.

\paragraph{Model Utilization.}
In terms of model selection, the ChatGPT family\cite{achiam2023gpt} dominates the landscape, being used in all 12 studies, as shown in Fig. \ref{fig:res}(b). Other models, such as Gemini (n = 3) and Perplexity (n = 2), appear less frequently, while models including Qwen\cite{bai2023qwen}, DeepSeek\cite{guo2025deepseek}, Gemma\cite{team2024gemma}, LLaMA\cite{touvron2023llama}, Grok, and Claude are each represented in only a single study.

This distribution indicates a strong reliance on a small set of widely accessible, general-purpose LLMs, with limited exploration of alternative or domain-specific models. The heavy concentration around the ChatGPT family may also reflect its accessibility and strong baseline performance, but it raises concerns about the diversity of model evaluation and the generalizability of findings across different architectures.

\paragraph{Data Sources and Application Scenarios.}

The majority of studies rely on patient-centered question sets, including FAQs and expert-curated questions (n = 9), as their primary data source, as shown in Fig. \ref{fig:res}(g-h). Other data types, such as patient education text generation samples (n = 3) and doctor--patient communication scenarios (n = 2), are less commonly used. Only a small number of studies incorporate structured patient data (n = 1), suggesting limited engagement with real-world, heterogeneous data sources.

Consistent with the data distribution, as shown in Fig. \ref{fig:res}(h), the most common application scenario is patient question answering and information retrieval (n = 6). This is followed by patient education content generation and dissemination (n = 4), and clinical workflow support tasks such as patient matching or counseling (n = 2). Fewer studies explore communication effectiveness (n = 1).

Overall, these findings suggest that current research primarily focuses on relatively controlled and well-defined tasks, such as question answering, while more complex, real-world applications remain underexplored. 

\paragraph{Evaluation Focus.}
Across the included studies (Fig. \ref{fig:res}(f)), evaluation metrics are primarily centered on accuracy or correctness (n = 8), followed by trustworthiness or reliability (n = 6). Other dimensions, such as readability and comprehensibility (n = 4), risk--benefit and safety considerations (n = 3), and communication quality or patient experience (n = 2), receive comparatively less attention. Scalability and system-level performance are also infrequently assessed (n = 2).

This distribution indicates that existing evaluations are largely focused on technical correctness, with relatively limited emphasis on patient-centered outcomes, such as usability, understanding, and real-world impact. Given the context of patient education, this imbalance highlights an important gap in current research and underscores the need for more holistic evaluation frameworks.

\section{Discussions}
\subsection{Main findings}
The application of LLMs to rare disease patient education and health communication represents a field still in its very early stage, one that only became technically feasible with the recent generation of large-scale, instruction-tuned language models. The contrast with the broader patient education literature is significant: a 2024 scoping review\cite{aydin2024large} identified 201 studies using LLMs for patient education in general medical domains, and a 2025 focused review of LLM-based generation of patient education materials\cite{alsammarraie2025use} identified 69 studies, while our review identified only 12 rare disease-specific literature in early 2026, among which only two explored content generation. Given the distinct and complex communication needs in rare diseases, this gap highlights a domain with substantial potential that remains largely underexplored.

Most existing studies adopt a relatively narrow evaluation framework, assessing how well off-the-shelf chatbots, primarily ChatGPT (GPT-4), but also Gemini, Claude, and Perplexity, respond to predefined disease-specific questions. Typically, these studies compile a set of questions, submit them to one or more models, and ask clinicians or domain experts to rate the responses generated by LLMs for accuracy and completeness. Across studies, ChatGPT shows reasonable to good accuracy (approximately 70–100\%) on rare disease-related questions, with other models showing broadly comparable performance. Notably, only one study moved beyond off-the-shelf querying to integrate external rare disease knowledge bases, using Orphanet to inform GPT-4 prompting and to adapt BioMistral-7B fine-tuned via low-rank adaptation (LoRA)\cite{hu2022lora}. The absence of retrieval-augmented or domain-adapted approaches across the main body of work is a significant gap, as general-purpose LLMs trained on web-scale resources are inherently limited in their coverage of rare conditions for which published literature could be sparse, outdated, or concentrated in highly specialized sources not well-represented in pretraining data.

The definition and selection of patient questions also varies considerably across studies and reflects important methodological choices with real implications for generalizability. Two studies restricted their question sets to purely clinical and disease-specific questions\cite{valentini2024artificial, perez2025enhancing}, while two others used real-world patient questions drawn from existing datasets\cite{tao2025can, weber2025medbot}, and the other six included a broader range spanning general disease information, diagnosis, prognosis, lifestyle management, and lived experience\cite{sevincc2025academic, henriques2025artificial, zhao2025automating, cilli2026chatgpt, jeon2025evaluating, van2025risks}. Only three studies explicitly grounded their question development in patient advocacy organization resources or online health information platforms\cite{henriques2025artificial, zhao2025automating, van2025risks}, suggesting that the majority of question sets reflect clinician assumptions about what patients ask rather than empirically documented patient information needs. 

Critically, patient-centered evaluation has not yet been systematically adopted in this field. While accuracy is assessed in ten of the twelve studies, fewer than half consider dimensions such as comprehensiveness or readability as additional dimensions, and only two studies considered empathy or performance in emotionally sensitive scenarios\cite{weber2025medbot, van2025risks}. This gap is further compounded by the choice of evaluators: most studies that used human raters relied exclusively on clinicians, medical experts, or information professionals, and only two engaged multiple stakeholders, including patient advocacy leaders or community representatives\cite{perez2025enhancing, van2025risks}. This absence of patients themselves as evaluators represents a fundamental blind spot in a study aimed at serving these populations.

An additional structural limitation that presents in most of the studies is the limited attention to multilingual communications. Among the twelve studies identified, nine were conducted exclusively in English, one was conducted exclusively in Korean\cite{jeon2025evaluating}, and only two incorporated additional languages — one generating educational podcast content in Spanish and German alongside English\cite{perez2025enhancing}, and one assessing comprehensive proficiency in both German and English\cite{weber2025medbot}. This is particularly problematic in the context of rare diseases, where the geographic and ethnic distribution of specific conditions frequently does not align with English-speaking populations. The familial Mediterranean fever study\cite{cilli2026chatgpt} is the only work to explicitly acknowledge this mismatch, noting that the exclusive use of English restricts the assessment to anglophone contexts, and that this design choice may substantially limit the generalizability of findings given that FMF predominantly affects Turkish-, Arabic-, and Hebrew-speaking populations. More broadly, the lack of multilingual evaluation raises concerns about the generalizability, equity, and real-world utility of LLM-based communication tools, especially given the importance of culturally and linguistically appropriate communication in patient-centered care.



 
\subsection{Future Research Priorities}
Addressing the limitations identified in the current literature requires a shift from proof-of-concept evaluations toward the development of robust, patient-centered, and clinically grounded LLM systems. First, future work should move beyond static question–answer benchmarks toward task-oriented and longitudinal evaluation frameworks\cite{zhu2024prompting} that reflect real-world communication scenarios. This includes multi-turn dialogue, evolving patient needs across the disease trajectory, and context-aware interactions that incorporate prior history, preferences, and uncertainty. Standardized evaluation protocols should extend beyond accuracy to include dimensions such as readability, usefulness, actionability, empathy, uncertainty communication, and alignment with patient values. Critically, patients and caregivers must be systematically involved as evaluators and co-design partners, together with clinicians, to ensure that evaluation metrics reflect real communication needs rather than proxy clinical judgments.

Second, there is a clear need to develop and rigorously assess domain-adapted and knowledge-grounded LLM approaches. Retrieval-augmented generation (RAG)\cite{zhan2025mmrag,zhan2026raicl} pipelines integrating curated rare disease resources (e.g., Orphanet, clinical guidelines, patient organization materials) represent a promising direction to improve factual accuracy and coverage for the large number of rare diseases. Fine-tuning or parameter-efficient adaptation strategies (e.g., LoRA\cite{hu2022lora}) on high-quality domain-specific corpora, including clinical notes and patient-generated data, may further enhance relevance and personalization. Future studies should compare these approaches systematically against off-the-shelf models and evaluate their performance in low-resource settings where data scarcity is intrinsic.

Third, research should explicitly address multilingual and cross-cultural communication\cite{luo2023cross}, which remains largely neglected despite its central importance in rare diseases. This includes evaluating LLM performance across languages, adapting models to culturally specific communication norms, and ensuring equitable access to high-quality information for non-English-speaking populations. Incorporating multilingual corpora and region-specific knowledge sources will be essential to improve generalizability and reduce disparities.

Fourth, future work should focus on integration into clinical and real-world workflows. Rather than evaluating LLMs in isolation, studies should examine how these tools can support communication between patients and healthcare providers, facilitate care coordination, and augment existing digital health systems such as patient portals. This includes assessing usability, trust, and adoption, as well as understanding how LLM outputs influence patient behavior, decision-making, and clinical outcomes.

Finally, ensuring safety, transparency, and ethical use is critical for deployment in rare disease contexts. Given the high stakes and inherent uncertainty, future research must address issues such as hallucination, calibration of uncertainty, provenance of information, and mechanisms for human oversight. Frameworks for responsible use should also consider data privacy, bias, and the risk of over-reliance on automated systems.

Together, these priorities highlight the need for a multidisciplinary research agenda that combines advances in natural language processing with clinical expertise, patient engagement, and health communication science to develop safe, effective, and equitable LLM-based solutions for rare disease patient communication.


\section{Conclusion}
This scoping review highlights that the application of LLMs in rare disease patient education and communication is still at an early and exploratory stage, with existing studies largely focusing on simplified question-answering tasks and accuracy-based evaluations using general-purpose models. While these approaches demonstrate promising potential in delivering reasonably accurate information, they fall short in addressing the complex, longitudinal, and patient-centered communication needs inherent to rare diseases, including aspects such as readability, empathy, cultural and linguistic appropriateness, and real-world usability. Moreover, the limited use of real-world data, insufficient involvement of patients in evaluation, and lack of multilingual considerations constrain the generalizability and practical impact of current research. Nevertheless, LLMs offer a compelling opportunity to bridge gaps in knowledge access, support personalized and adaptive communication, and enhance patient empowerment in rare disease contexts. Realizing this potential will require a shift toward domain-adapted, knowledge-grounded approaches, comprehensive and patient-centered evaluation frameworks, and careful integration into clinical workflows, with strong attention to safety, transparency, and equity.


\bibliographystyle{IEEEtran}
\bibliography{0_reference}

\end{document}